\documentclass[10pt,twocolumn,letterpaper]{article}

\usepackage[pagenumbers]{cvpr} 

\definecolor{cvprblue}{rgb}{0.21,0.49,0.74}
\usepackage[pagebackref,breaklinks,colorlinks,allcolors=cvprblue]{hyperref}
\usepackage{xspace}
\usepackage{array}
\newcommand{\name}{SWiT-4D\xspace}

\title{SWiT-4D: Sliding-Window Transformer for Lossless and Parameter-Free Temporal 4D Generation}

\author{
Kehong Gong$^{*,1}$, Zhengyu Wen$^{*,2}$, Mingxi Xu$^{*,2}$, Weixia He$^{2}$, Qi Wang$^{2}$, 
Ning Zhang$^{2}$, Zhengyu Li$^{2}$,\\ Chenbin Li$^{2}$, Dongze Lian$^{2}$, Wei Zhao$^{2}$, Xiaoyu He$^{2}$, Mingyuan Zhang$^{\dagger,2}$ \\
$^{1}$Huawei Technologies Co., Ltd., $^{2}$Huawei Central Media Technology Institute \\
{\small $^{*}$ Equal Contributions, $^{\dagger}$ Corresponding Author} \\
}

\begin{document}

\maketitle


\begin{abstract}

\vspace{-5pt}

Despite significant progress in 4D content generation, the conversion of monocular videos into high-quality animated 3D assets with explicit 4D meshes remains considerable challenging. The scarcity of large-scale, naturally captured 4D mesh datasets exacerbates the difficulty of learning generalizable video-to-4D models from scratch in a purely data-driven manner.
Fortunately, the substantial progress in image-to-3D, supported by extensive datasets, offers powerful prior foundation models.
To better leverage these models while minimizing reliance on 4D supervision, we introduce a novel method, \name—a Sliding-Window Transformer for lossless, parameter-free temporal 4D mesh generation. \name can be seamlessly integrated with any Diffusion Transformer (DiT)-based image-to-3D generator, augmenting it with spatial-temporal modeling capabilities across video frames while preserving its original single-image forward process, which enables 4D mesh reconstruction from videos of arbitrary length. 
To recover global translation, we further introduce an optimization-based trajectory module tailored for static-camera monocular videos.
Remarkably, \name demonstrates strong data efficiency: with only a single short ($<$10s) video for fine-tuning, our model attains high-fidelity geometry and stable temporal consistency, highlighting its practical deployability even under extremely scarce 4D supervision.
Comprehensive experiments on both in-domain zoo-test sets and challenging out-of-domain benchmarks (c4d, Objaverse, and in-the-wild videos) show that our method consistently outperforms existing baselines in temporal smoothness, underscoring the practical merits of the proposed framework.  Project page: \url{https://animotionlab.github.io/SWIT4D/}

\end{abstract}    
\section{Introduction}
\label{sec:intro}

Video-conditioned 4D generation aims to reconstruct dynamic 3D objects and their textures from input videos.
With the rapid progress of generative modeling for images~\citep{flux2024}, videos~\citep{wan2025}, and 3D geometry~\citep{hunyuan3d22025tencent,xiang2024trellis,li2025triposg},
this task has recently gained increasing attention in both computer vision and graphics communities.
However, the leap from static 3D asset reconstruction to dynamic 4D mesh generation introduces substantial new challenges:
(i) the lack of large-scale real-world 4D mesh datasets,
(ii) the necessity of temporal coherence across arbitrary video lengths,
(iii) and the need to preserve the strong generalization ability of state-of-the-art image-to-3D models.
While image-to-3D diffusion transformers have achieved remarkable single-view 3D reconstruction quality~\citep{li2025triposg,li2025step1x},
they remain limited to static scenes and cannot reason across time.

Early 4D generation methods rely on Score Distillation Sampling (SDS)~\citep{poole2022dreamfusion} to optimize dynamic geometry~\citep{singer2023text4d},
but suffer from fragility, high computational cost, and slow convergence.
Two-stage approaches~\citep{zhang20244diffusion,wu2024cat4d,wang20254realvideov2fusedviewtimeattention}
address these issues by first generating multi-view videos and then reconstructing 3D or 4D shapes,
but reconstruction errors and view inconsistencies accumulate during generation.
More recent frameworks~\citep{chen2025v2m4,zhang2025gaussian}
attempt to extend pretrained 3D generators for 4D tasks,
yet they either rely on optimization-based pipelines or introduce new deformation fields and networks,
leading to increased complexity and weaker generalization.
Due to the scarcity of 4D data, training a video-to-4D model entirely from scratch is highly challenging, and even when leveraging pretrained image-to-3D priors, naive finetuning on limited 4D supervision often degrades the original generalizability. The recent V2M4~\citep{chen2025v2m4} attempts to alleviate this issue by first generating a 3D mesh for each frame and then enforcing temporal stability via post-hoc optimization. While this strategy reduces the dependence on 4D data, it also prevents the model from fully exploiting high-quality 4D supervision to further improve reconstruction quality.
In summary, current approaches either rely on expensive optimization or introduce new trainable modules, which inevitably compromise generalization and efficiency.
This motivates a minimalistic and lossless extension for 4D mesh generation that preserves the advantages of state-of-the-art image-to-3D backbones.

Fortunately, the substantial progress in image-to-3D, supported by extensive datasets, offers powerful prior foundation models. Therefore, a desirable approach for video-to-4D is to inherit the pre-trained weights of a powerful image-to-3D model, thereby preserving its strong generalization capabilities while enforcing temporal coherence for video-to-4D generation.
To this end, we introduce a sliding window mechanism to process the input sequence. Such a deliberate design elegantly fulfills the aforementioned objectives with the following advantages:
i) As substantiated in our methodology, our design inherently and losslessly retains the pre-trained weights for image-to-3D generation, enabling seamless fine-tuning. In essence, single-frame reconstruction is a special case of our generalized framework, and conversely, our method acts as a temporal extension of the single-image model. This allows for highly efficient parameter utilization and, crucially, retains the powerful generalization capabilities of the original model, leading to superior performance on out-of-domain (OOD) data.
ii) Owing to the position-agnostic nature of the sliding window rather than the insertion of a temporal block, our framework can be naturally applied to long 4D sequences. As subsequently demonstrated in our experiments, it maintains high-fidelity and temporally consistent results across extended 4D sequences.
iii) Training Efficiency: The framework converges rapidly within only 4 hours of training for a model with 1.5B parameters, representing a computationally efficient solution.

In addition, to recover the global translation of the mesh in the world coordinate system, we introduce an optimization-based trajectory prediction module tailored for static-camera monocular videos. Specifically, we extract foreground regions from the rendered mesh and the input RGB frames, and optimize the global pose by maximizing the IoU between these masks. This yields accurate world-space trajectories, making our generated 4D meshes more readily usable in downstream applications.

In summary, we propose a new video-to-4D framework with the following advantages:
\begin{enumerate}
\item \textbf{Strong generalization.} Our model is built on a powerful image-to-3D generator, and its single-frame inference is \emph{exactly} identical to the pretrained backbone, thus maximally preserving the original generalization ability.
\item \textbf{Efficient training.} The sliding-window temporal mechanism, instead of inserting extra temporal blocks, enables fast convergence and a smooth transition from a single-frame model to a temporally consistent video-to-4D model with a few training data.
\item \textbf{Accurate global trajectories.} Our optimization-based global trajectory prediction module complements the data-driven components and provides accurate global motion estimates for the generated 4D sequences.
\end{enumerate}

\section{Related Work}
\label{sec:related}

\begin{figure*}[t]
  \centering
  \includegraphics[width=0.95\textwidth]{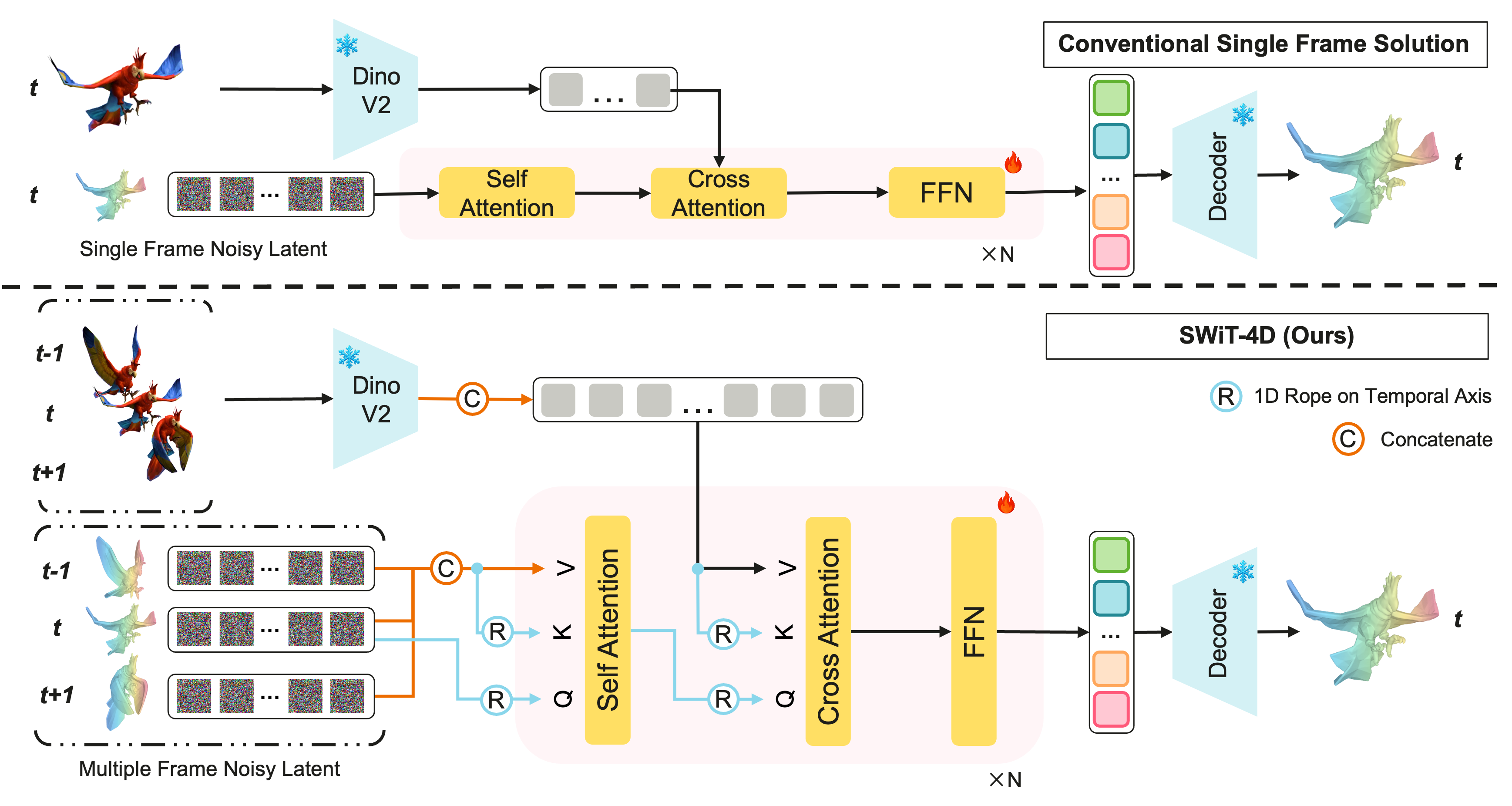} 

  \caption{SWiT-4D, a parameter-free temporal extension to image-to-3D diffusion transformers.
Top: Conventional single-frame 3D generation. A shape VAE encodes each 3D mesh into latent space, the diffusion transformer performs denoising, and the shape VAE decodes the latent back to 3D geometry—without any temporal reasoning.
Bottom: Our method introduces temporal modeling losslessly through a sliding-window mechanism applied to both self- and cross-attention. A 1D rotary positional encoding (1D-RoPE) encodes temporal phase, ensuring identical behavior to the single-frame model when $W{=}0$, while enabling temporal residual learning when $W{>}0$. This design allows coherent motion perception and temporally consistent 4D generation without adding new parameters or supervision.}
  \label{fig:pipeline}
\vspace{-10pt}
\end{figure*}

\subsection{Image-to-3D Generation}

Single-image 3D generation has seen rapid progress thanks to large-scale 3D datasets and scalable transformer architectures. 
Some methods follow a two-stage “2D lifting” pipeline: given one/few views, a multi-view generator first synthesizes novel views, then a reconstructor lifts them into 3D—e.g., One-2-3-45/++~\citep{liu2023one2345,liu2023one2345pp,DBLP:conf/iccv/LiuWHTZV23} (fixed-view diffusion + SDF), Wonder3D~\citep{long2024wonder3d} (RGB+normal diffusion with normal fusion), SyncDreamer~\citep{liu2024syncdreamer} (single-view multi-view synthesis), and the tighter couplings Cycle3D~\citep{tang2025cycle3d} and InstantMesh~\citep{xu2024instantmesh}, which improve fidelity and robustness. Complementary to this, native 3D generative models learn a latent 3D space with diffusion/flow Transformers: TRELLIS~\citep{xiang2024trellis} (structured 3D latent, multi-format decoding), Hunyuan3D-2.0~\citep{hunyuan3d22025tencent} (flow-based shape prior + separate texture), Step1X-3D~\citep{li2025step1x} (VAE–DiT shape + diffusion texture), and TripoSG~\citep{li2025triposg} (rectified-flow over structured 3D tokens for clean meshes). These image-conditioned backbones provide strong priors for static geometry but lack explicit temporal modeling, leaving a gap for video-conditioned 4D reconstruction—the focus of our work.

\subsection{Video-to-4D Reconstruction and Generation}

Due to the scarcity of 4D assets, many methods reconstruct 4D content from monocular or multi-view videos by leveraging priors from 2D generative models~\citep{jiang2024consistent4d,zeng2024stag4d,zhu2025ar4d,yang2024diffusion2,sun2024eg4d,jiang2024animate3d,zhang20244diffusion,liang2024diffusion4d,xie2024sv4d,ren2024l4gm,li2024dreammesh4d}. A complementary direction adapts pretrained 3D generators to video-conditioned 4D prediction. V2M4~\citep{chen2025v2m4} performs per-frame image-to-3D generation followed by post-hoc temporal optimization, while GVFD~\citep{zhang2025gaussian} augments 3D diffusion transformers with deformation fields or Gaussian particles, introducing extra networks and supervision. Concurrent to our work, ShapeGen4D~\citep{ShapeGen4D} adds dedicated spatio–temporal layers and losses on top of image-to-3D diffusion models for 4D mesh generation. In contrast, SWiT-4D augments the self- and cross-attention of a pretrained image-to-3D DiT with a \emph{parameter-free} sliding-window temporal mechanism based on 1D RoPE, preserving single-frame behavior and enabling efficient temporal residual learning for multiple frames without changing the backbone architecture or adding extra trainable parameters.
\section{Method}
\label{sec:method}

\subsection{Overview}
To address the scarcity of 4D supervision, we ground our method in strong image-to-3D priors and fine-tune on a few high-quality 4D data, injecting temporal consistency while preserving the original generalization. Consequently, we aim to extend pretrained image-to-3D diffusion transformers (e.g., TripoSG~\citep{li2025triposg}) into temporally consistent 4D generators \emph{without} introducing additional parameters, supervision, or losses. To this end, we propose \textbf{\name}—a \textbf{S}liding-\textbf{Wi}ndow \textbf{T}ransformer for lossless temporal \textbf{4D} generation. In Section~\ref{sec:prelim} we will first introduce the widely used DiT-based image-to-3D generation model, which is the foundation of our proposed \name. Section~\ref{sec:attention} details how we augment standard self/cross-attention with a sliding-window temporal mechanism; Section~\ref{sec:proof} proves that our design maximally preserves the priors learned by the pretrained image-to-3D backbone. The training and inference pipelines are presented in Sections~\ref{sec:training} and \ref{sec:inference}. Finally, to handle global, world-space motion, we introduce a trajectory estimation module, described in Section~\ref{sec:trajectory}.

\subsection{Preliminaries: Image-Conditioned 3D Generative Backbones}
\label{sec:prelim}

A mainstream image-to-3D paradigm~\cite{li2025triposg,li2025step1x,hunyuan3d22025tencent} trains an image-conditioned DiT/flow prior on a latent 3D shape code $x$ and decodes it to a mesh $\hat M$; conditioning images $I$ are mapped to tokens $C=f_{\mathrm{img}}(I)$ that are injected via cross-attention. The backbone is a Transformer operating on the set of latent 3D tokens $x\in\mathbb{R}^{L\times d}$ together with image tokens $C\in\mathbb{R}^{N_c\times d_c}$: each block contains (i) self-attention on $x$ to model long-range geometric/topological correlations within the single-frame 3D token set, (ii) cross-attention from $x$ (queries) to $C$ (keys/values) to inject image cues as conditions guiding 3D synthesis, and (iii) a position-wise feed-forward sublayer. The prior is learned with the (rectified) flow-matching objective:
\begin{equation}
\label{eq:prelim-fm}
\mathcal{L}_{\mathrm{FM}}
=
\mathbb{E}_{x,\epsilon,s}\!\left[\;\big\|u_\theta\!\big(h_s(x,C),\,s\big)-v^\star(x,\epsilon,s)\big\|_2^2\;\right],
\end{equation}
where $h_s(x,C)$ is the DiT hidden state at flow time $s$ given latent tokens $x$ and image tokens $C$; $u_\theta$ is the velocity field predicted by the backbone with parameters $\theta$; $v^\star(x,\epsilon,s)$ is the target velocity defined by the chosen (rectified-)flow schedule (e.g., $x_1{-}x_0$), with $\epsilon$ denoting any auxiliary randomness. At inference the latent is integrated by the flow and then decoded:
\begin{equation}
\label{eq:prelim-ode}
\dot x_s = u_\theta\!\big(h_s(x_s,C),\,s\big),
\end{equation}
\begin{equation}
\label{eq:prelim-decode}
\hat M = D_g(\hat x),
\end{equation}
where $\dot x_s$ denotes the derivative w.r.t.\ flow time $s$, $\hat x$ is the terminal latent obtained by integrating Equation \eqref{eq:prelim-ode}, and $D_g$ is a geometry decoder that outputs a mesh $\hat M=(\hat V,\hat F)$. Within this shared paradigm, \textit{TripoSG}~\citep{li2025triposg} trains a rectified-flow DiT over structured 3D tokens $x\!\in\!\mathbb{R}^{L\times d}$ with a geometry tokenizer $E_g$ and mesh decoder $D_g$ (so $x_1=E_g(y)$ and $\hat M=D_g(\hat x)$), optionally adding geometric regularizers on decoded shapes; \textit{Step1X-3D}~\citep{li2025step1x} factorizes assets into a shape latent $x_s$ trained by Equation \eqref{eq:prelim-fm} via a 3D VAE $(E_s,D_s)$ and a texture/material latent $x_t$ decoded by $D_t$ and supervised in UV/render space with a differentiable renderer; \textit{Hunyuan3D-2.0}~\citep{hunyuan3d22025tencent} follows a two-stage design with a large flow-based shape prior trained by \eqref{eq:prelim-fm} and a separate PBR head, optimized by rendering on $\hat M$. All three are image-conditioned DiT/flow backbones of the form \eqref{eq:prelim-fm}–\eqref{eq:prelim-decode}; in this paper we instantiate and evaluate \name primarily on TripoSG, and later (Section \ref{sec:attention}) replace the single-frame state $h_s(x,C)$ by its parameter-free sliding-window temporal counterpart.

\subsection{Sliding-Window Temporal Attention}
\label{sec:attention}

To strengthen temporal modeling without introducing any new parameters, we augment (i) the self-attention in a subset of layers with a sliding-window, 1D-RoPE–based temporal mechanism so the network can capture motion dynamics, and (ii) the cross-attention in another subset so the conditioning pathway can aggregate temporally adjacent visual evidence, providing more clues about temporal motion. 
This minimal and unified approach preserves the original model’s generalization, while granting powerful temporal reasoning to both geometry and image conditioning.

We encode temporal relations by applying a 1D rotary positional encoding (1D-RoPE) along time to queries/keys at every timestamp. Specifically, let $\mathbf{Q},\mathbf{K},\mathbf{V}\in\mathbb{R}^{T\times N\times D}$ denote per-frame token sets (time $T$, tokens-per-frame $N$, width $D$), with frame-$t$ slices $\mathbf{Q}_t,\mathbf{K}_t,\mathbf{V}_t\in\mathbb{R}^{N\times D}$ and token vectors $q_{t,i},k_{t,i},v_{t,i}\in\mathbb{R}^D$; we apply $R_t\in\mathbb{R}^{D\times D}$ to all tokens at time $t$ as $q'_{t,i}=R_t q_{t,i}$ and $k'_{t,i}=R_t k_{t,i}$, where $R_t$ is a block-diagonal orthogonal rotation defined by sinusoidal phases. This temporal RoPE yields:
\begin{enumerate}[label=(\roman*),leftmargin=*,itemsep=0pt,topsep=2pt]
\item \emph{Relative-time sensitivity.} For any offset $\Delta$, the similarity depends only on the temporal \emph{difference}: $(q'_{t,i})^\top k'_{t+\Delta,j}=q_{t,i}^\top(R_t^\top R_{t+\Delta})k_{t+\Delta,j}=q_{t,i}^\top(R_{\Delta}k_{t+\Delta,j})$, yielding shift-equivariant temporal scoring and invariance to absolute timestamps. 
\item \emph{Single-frame equivalence.} When the temporal window collapses to the current frame ($W{=}0$), $R_t^\top R_t=I$ and the operation reduces exactly to the original image-to-3D attention, guaranteeing lossless prior preservation. We will prove it in the later section.
\item \emph{Length extrapolation (train short, infer long).} Because $R_\Delta$ is sinusoidal (hence well-defined for unseen $t,\Delta$) and the sliding-window attention shares parameters across time, a model trained on short clips can be applied to arbitrarily long sequences at test time with memory/compute growing only linearly on total sequence length. 
\end{enumerate}
Given a clip $\{x_{t-W},\ldots,x_t,\ldots,x_{t+W}\}$, the temporal context for frame $t$ is the index set $\Omega_t=\{\tau\mid |\tau-t|\le W\}$; for a query token $q'_{t,i}$ we attend over \emph{all} tokens from all frames in the window, i.e., $\{k'_{\tau,j},v_{\tau,j}\mid \tau\in\Omega_t,\, j\in[1..N]\}$; stacking keys/values in the window as $K'_{\Omega_t}\in\mathbb{R}^{(2W+1)N\times D}$ and $V_{\Omega_t}\in\mathbb{R}^{(2W+1)N\times D}$, the token-wise attention is 
\begin{equation}
\mathrm{Attn}(q'_{t,i})=\mathrm{Softmax}\!\Big(\frac{q'_{t,i}(K'_{\Omega_t})^\top}{\sqrt{D}}\Big)\,V_{\Omega_t},
\label{eq:sw-token}
\end{equation}
and the frame-wise form is 
\begin{equation}
\begin{split}
\mathrm{Attn}(\mathbf{X}_t)=\mathrm{Softmax}\!\Big(\frac{\mathbf{Q}'_t (K'_{\Omega_t})^\top}{\sqrt{D}}\Big)\,V_{\Omega_t}, \\
\mathbf{Q}'_t=\mathbf{Q}_t R_t^\top, K'_{\Omega_t}=\mathrm{stack}_{\tau\in\Omega_t}(\mathbf{K}_\tau R_\tau^\top),
\label{eq:sw-attn}
\end{split}
\end{equation}
which introduces temporal reasoning \emph{without adding parameters}; for fixed $N,D$, the number of keys per query grows from $N$ to $(2W{+}1)N$, so memory/compute scale linearly with $W$, and at inference we maintain a rolling KV cache over $\Omega_t$ to slide across arbitrarily long sequences with fixed model size.

\subsection{Lossless Temporal Property (Proof)}
\label{sec:proof}
RoPE is applied as $q'_{t,i}=R_t q_{t,i}$ and $k'_{t,j}=R_t k_{t,j}$ with $R_t^\top R_t=I$. Therefore, when $W{=}0$ (i.e., $\Omega_t=\{t\}$), all keys/values come from the same frame and the score between any token pair is preserved: 
\begin{equation}
s'_{ij} \!=\! (q'_{t,i})^\top k'_{t,j} \!=\! q_{t,i}^\top R_t^\top R_t k_{t,j} \!=\! q_{t,i}^\top k_{t,j} \!=\! s_{ij}.
\end{equation}
This suggests that the attention weights are identical, $\alpha'_{ij} \!=\! \mathrm{Softmax}_j(s'_{ij}/\sqrt{D}) \!=\! \mathrm{Softmax}_j(s_{ij}/\sqrt{D}) \!=\! \alpha_{ij}$, and the outputs coincide token-wise, $o'_{t,i} \!=\! \sum_j \alpha'_{ij} v_{t,j} \!=\! \sum_j \alpha_{ij} v_{t,j} \!=\! o_{t,i}$, which in matrix form yields $\mathrm{Attn}(\mathbf{Q}'_t,\mathbf{K}'_t,\mathbf{V}_t)\!=\!\mathrm{Attn}(\mathbf{Q}_t,\mathbf{K}_t,\mathbf{V}_t)$. Hence, \name is \textbf{lossless} at $W{=}0$, exactly preserving the single-frame behavior of the pretrained image-to-3D backbone (including identical masking, bias, and normalization paths), and at initialization—when weights are inherited, our \name and the used pretrained image-to-3D model are functionally equivalent for any single-frame input.

\paragraph{Temporal residual learning.}
When $W{>}0$, a query $q_{t,i}$ attends to all tokens in the window $\Omega_t$ and cross-frame scores incorporate the \emph{relative} rotation $R_t^\top R_{t+\Delta}$: $({q'_{t,i}})^\top k'_{t+\Delta,j}\!=\!q_{t,i}^\top (R_t^\top R_{t+\Delta})k_{t+\Delta,j}\!=\!q_{t,i}^\top (R_{\Delta}k_{t+\Delta,j})$, where $R_\Delta$ depends only on the offset $\Delta$. Decomposing the softmax over $\Omega_t$ gives an intra-frame term (the preserved static prior) plus inter-frame corrections (temporal residuals), so the token output can be written schematically as
\begin{equation}
\mathrm{Attn}_{t,i}=\underbrace{\mathrm{Attn}(q_{t,i},\mathbf{K}_t)}_{\text{static prior}}+\sum_{\Delta\neq 0}\underbrace{\mathrm{Attn}(q_{t,i},\mathbf{K}_{t+\Delta})}_{\text{temporal residual}},
\end{equation}
with the same value vectors $\mathbf{V}$ and no new parameters. Practically, the residual magnitude and range are controlled by the window half-width $W$ (linear KV growth from $N$ to $(2W{+}1)N$ keys per query), enabling \name to inject motion cues that improve temporal coherence over arbitrarily long sequences while leaving the single-frame pathway unchanged.

\subsection{Training Objective}
\label{sec:training}
We preserve the original flow-matching training objective~\citep{li2025triposg}, 
which aligns model-predicted velocity fields $u_\theta$ with target flows $v^\star$:
\begin{equation}
\mathcal{L}_{\mathrm{FM}} =
\mathbb{E}_{x,\epsilon,s,t}
\big[
\|u_\theta(h_s(X_{t-W:t+W}), s) - v^\star(x, \epsilon, s)\|_2^2
\big].
\end{equation}
No additional temporal or contrastive loss is required. 
When $W{=}0$, this formulation exactly reduces to the original single-frame training.
When $W{>}0$, the model learns to incorporate temporal context within the same objective, 
emerging as temporal residual refinement without altering the training pipeline.

\subsection{Inference and Properties}
\label{sec:inference}
During inference, the sliding-window mechanism allows the model to process videos of arbitrary length by shifting the temporal window across frames, while keeping computation and memory bounded by $O(W)$.
Because \name introduces no additional parameters, 
it inherits the pretrained 3D model’s generalization ability and efficiency.
The lossless initialization ensures consistent geometry and texture at $W{=}0$, 
while fine-tuning with $W{>}0$ introduces temporal consistency and motion awareness.
The same design can be applied to any image-to-3D backbone 
(e.g., TripoSG~\citep{li2025triposg}, Step1X-3D~\citep{li2025step1x}, Hunyuan3D~\citep{hunyuan3d22025tencent}),
providing a unified and parameter-free pathway toward temporally coherent 4D generation.

\subsection{Trajectory Optimization}
\label{sec:trajectory}

To recover the motion trajectory of the generated meshes, we optimize their poses across all video frames in the world coordinate system. Since the scale and rotation of each mesh are already well estimated by our generation model, the optimization mainly refines the translations and camera parameters. We use a differentiable renderer to project each mesh onto the image plane and compare the rendered masks with the ground-truth masks extracted from the input video frames. This mask-based loss provides differentiable supervision, allowing gradients to pass through the rendering process and jointly refine both mesh translations and camera parameters.

Formally, let $M_i$ denote the mesh at frame $i$, $\theta_i$ its translation, and $c_i$ the camera parameters. The rendered mask is $R(M_i, \theta_i, c_i)$, and the ground-truth mask is $\hat{M}_i$. We define the mask loss as:
\begin{equation}
\mathcal{L}_{mask} =
\begin{aligned}
& \lambda_1 \mathcal{L}_{\text{BCE}}\big(R(M_i, \theta_i, c_i), \hat{M}_i\big) \\
& + \lambda_2 \mathcal{L}_{\text{Dice}}\big(R(M_i, \theta_i, c_i), \hat{M}_i\big)
\end{aligned}
\end{equation}
where $\lambda_1$ and $\lambda_2$ are weighting factors, and $\mathcal{L}_{\text{BCE}}$ refers to the \emph{Binary Cross-Entropy} loss, while $\mathcal{L}_{\text{Dice}}$ refers to the \emph{Dice Similarity Coefficient} loss.

To handle initial non-overlapping masks, we introduce a mask center loss. Let $(x_i, y_i)$ and $(\hat{x}_i, \hat{y}_i)$ denote the normalized centroid coordinates of the rendered and ground-truth masks at frame $i$. The center loss is:
\begin{equation}
\mathcal{L}_{center} = \frac{1}{N} \sum_{i=1}^{N} \big((x_i - \hat{x}_i)^2 + (y_i - \hat{y}_i)^2\big),
\end{equation}
where $N$ is the number of frames.

Finally, the overall loss with conditional weighting can be written using the aligned environment to prevent line overflow:
\begin{equation}
\mathcal{L} =
\begin{cases}
\begin{aligned}
& \epsilon \cdot \mathcal{L}_{\text{Dice}} + \zeta \cdot \mathcal{L}_{center} 
\end{aligned} 
& \text{if } \mathcal{L}_{\text{Dice}} > 0.999, \\[2.0ex]
\begin{aligned}
& \alpha \cdot \mathcal{L}_{\text{BCE}} + \beta \cdot \mathcal{L}_{\text{Dice}} \\
& + \gamma \cdot \mathcal{L}_{L1} + \delta \cdot \mathcal{L}_{center}
\end{aligned} 
& \text{otherwise.}
\end{cases}
\end{equation}

Using this adaptive loss design, the optimization produces a mesh sequence with accurate and temporally smooth trajectories that remain consistent with the observed video content.
\section{Experiments}
\label{sec:experiments}

\begin{table*}[]
\centering
\small
\begin{tabular}{lcccccccc}
\toprule
& \multicolumn{4}{c}{\textbf{Frame-Level Metrics}} & \multicolumn{4}{c}{\textbf{Temporal Metrics}} \\
\cmidrule(lr){2-5} \cmidrule(lr){6-9}
\textbf{Method} &
\textbf{CD} $\downarrow$ &
\textbf{F-score} $\uparrow$ &
\textbf{Precision} $\uparrow$ &
\textbf{Recall} $\uparrow$ &
\textbf{$\Delta$CD} $\downarrow$ &
\textbf{FE Cos} $\uparrow$ &
\textbf{Feat. DTW} $\downarrow$ &
\textbf{Occ. KL} $\downarrow$ \\
\midrule
L4GM & 0.0809 & 0.2158 & 0.1911 & 0.2646 & 0.0051 & 0.9641 & 3020.38 & 0.4298 \\
Genzoo & 0.1029 & 0.1643 & 0.1766 & 0.1581 & 0.0271 & 0.9737 & 2620.12 & 3.7381 \\
GVFD & 0.1406 & 0.1263 & 0.1263 & 0.1351 & 0.0159 & 0.9704 & 2224.90 & 1.6029 \\
TripoSG & 0.0863 & 0.2112 & 0.2103 & 0.2192 & 0.0437 & 0.9901 & 1714.63 & 2.4891 \\

\hline
Ours-1shot & 0.0594 & 0.2889 & 0.2835 & 0.2997 &
0.0047 & 0.9932 & 1598.41 & 0.3545 \\
\textbf{Ours} & \textbf{0.0338} & \textbf{0.4992} & \textbf{0.5031} & \textbf{0.4985} & \textbf{0.0027} & \textbf{0.9989} & \textbf{976.58} & \textbf{0.2312} \\
\bottomrule
\end{tabular}
\vspace{-5pt}
\caption{Quantitative comparison of methods on frame-level and temporal metrics on Truebones Zoo test dataset.}
\vspace{-5pt}
\label{tab:metrics-comparison-zoo}
\end{table*}

\begin{table*}[]
\centering
\small
\begin{tabular}{lcccccccc}
\toprule
& \multicolumn{4}{c}{\textbf{Frame-Level Metrics}} & \multicolumn{4}{c}{\textbf{Temporal Metrics}} \\
\cmidrule(lr){2-5} \cmidrule(lr){6-9}
\textbf{Method} &
\textbf{CD} $\downarrow$ &
\textbf{F-score} $\uparrow$ &
\textbf{Precision} $\uparrow$ &
\textbf{Recall} $\uparrow$ &
\textbf{$\Delta$CD} $\downarrow$ &
\textbf{FE Cos} $\uparrow$ &
\textbf{Feat. DTW} $\downarrow$ &
\textbf{Occ. KL} $\downarrow$ \\
\midrule
L4GM & 
0.1107 & 0.1806 & 0.2125 & 0.1652 &
0.0170 & 0.9830 & 2029.13 & 1.9931 \\

Genzoo & 
0.2182 & 0.0313 & 0.0389 & 0.0312 &
0.0360 & 0.9832 & 1601.70 & 4.0621 \\

GVFD & 
0.1157 & 0.1546 & 0.1515 & 0.1619 &
0.0231 & 0.9783 & 1961.06 & 2.8180 \\

TripoSG & 
0.1507 & 0.1112 & 0.1176 & 0.1114 &
0.0494 & 0.9911 & 1589.03 & 3.6063 \\

(+latent share) &
0.0958 & 0.1863 & 0.2050 & 0.1735 &
0.0271 & 0.9915 & \textbf{1587.79} & 2.0455 \\

(+latent smooth) &
0.0944 & 0.1925 & 0.2113 & 0.1797 &
0.0134 & \textbf{0.9918} & 1594.86 & 0.8864 \\

\hline

Ours-1shot & 
0.0789 & 0.1933 & 0.1934 & 0.1989 &
0.0086 & \textbf{0.9918} & 2136.81 & 0.7653 \\

Ours & 
\textbf{0.0786} & \textbf{0.2253} & \textbf{0.2293} & \textbf{0.2274} &
\textbf{0.0065} & 0.9911 & 1729.46 & \textbf{0.5616} \\

\bottomrule
\end{tabular}
\vspace{-5pt}
\caption{Quantitative comparison of methods on frame-level and temporal metrics on Objaverse test dataset. }
\vspace{-5pt}
\label{tab:objaverse-main}
\end{table*}


\subsection{Datasets and Experimental Setup}
\label{sec:datasets}

We conducted comprehensive experiments on the curated \textbf{Truebones Zoo} dataset, which contains 1,038 animal motion sequences with a total of 104,715 frames, covering a wide range of species and diverse motion patterns.

We consider two training settings to assess both data efficiency and large-scale learning:
(i) \textbf{One-shot LoRA Fine-tuning.}  
We selected three representative species: {Alligator}, {Horse} and {Parrot}. For each species, a single sequence of $\sim$150 frames is used for fine-tuning, while five additional sequences are sampled as the test set. This setup evaluates the adaptability of our method under extremely limited supervision.
(ii) \textbf{Full-scale Fine-tuning.}  
To evaluate the effect of large-scale multi-species learning, we construct a full training set containing \textbf{978 sequences} from Truebones Zoo, spanning diverse shapes, motions, and sequence lengths. The remaining \textbf{60 sequences} are held out as the test set.

To further assess the generalization capability of our model, we also evaluate both the one-shot and full-scale variants on several external datasets:  
(i) \textbf{Objaverse} (20 randomly sampled motion sequences),  
(ii) \textbf{Consistent4D}, and  
(iii) challenging \textbf{in-the-wild} videos.  
These additional test sets include assets and motions unseen during training, allowing for evaluation of both in-domain and out-of-domain generalization across novel shapes, topologies, and motion types.

\subsection{Evaluation Metrics}

We evaluate the results in both single-frame quality and temporal consistency. For single-frame quality, we use the following frame-level metrics, following~\cite{Zhang:2023:VecSet}:
\begin{itemize}
\item \textbf{Chamfer Distance (CD):} Measures geometric reconstruction accuracy.
\item \textbf{F-Score, Precision, Recall:} Evaluate mesh overlap at varying thresholds.
\end{itemize}

For evaluating temporal consistency, we introduce the following temporal metrics, with full computation details provided in the Supplementary Material:
\begin{itemize}

\item \textbf{Feature Cosine, Feature DTW.}  
These metrics quantify temporal alignment and motion coherence by comparing the evolution of point cloud features extracted from each mesh frame. We adopt PointNet++~\cite{DBLP:conf/nips/QiYSG17} as the feature encoder and compute cosine similarity and Dynamic Time Warping (DTW) distances across the feature trajectories.

\item \textbf{$\Delta$CD.}  
We compute the Chamfer Distance (CD) between consecutive frames for both the generated sequence and the ground truth. The absolute difference between these per-frame CDs measures discrepancies in the motion pattern, indicating how similarly the meshes evolve over time.

\item \textbf{Occupancy KL.}  
Analogous to $\Delta$CD, we compute the Kullback–Leibler (KL) divergence between voxel occupancy distributions of consecutive frames, and then take the absolute differences between generated and ground-truth values. This metric captures differences in the temporal evolution of volumetric occupancy, providing another perspective on motion similarity.

\end{itemize}

Arrows ($\uparrow$/$\downarrow$) indicate if higher/lower values are better.

\begin{figure*}[!t]
    \centering
    \includegraphics[width=\textwidth]{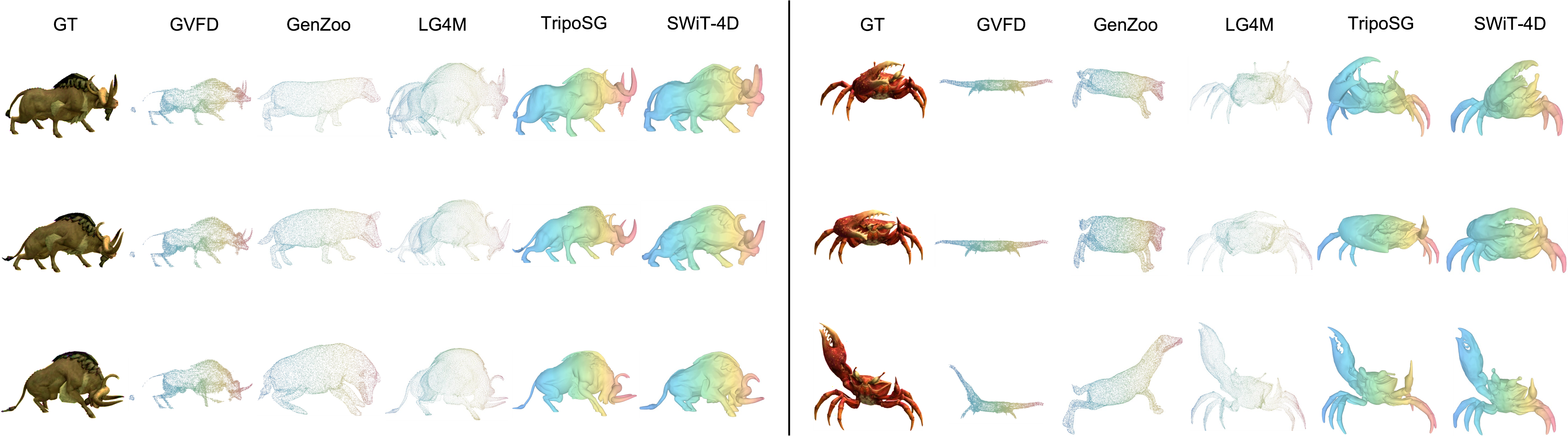}
    \caption{
    \textbf{Comparison with baselines on the Truebones Zoo test set.}
    Our method produces more temporally coherent poses, fewer skeletal artifacts, and more consistent motion semantics across the full sequence.
    }
    \label{fig:zoo-compare}
\vspace{-5pt}
\end{figure*}

\begin{figure*}[!t]
    \centering
    \includegraphics[width=\textwidth]{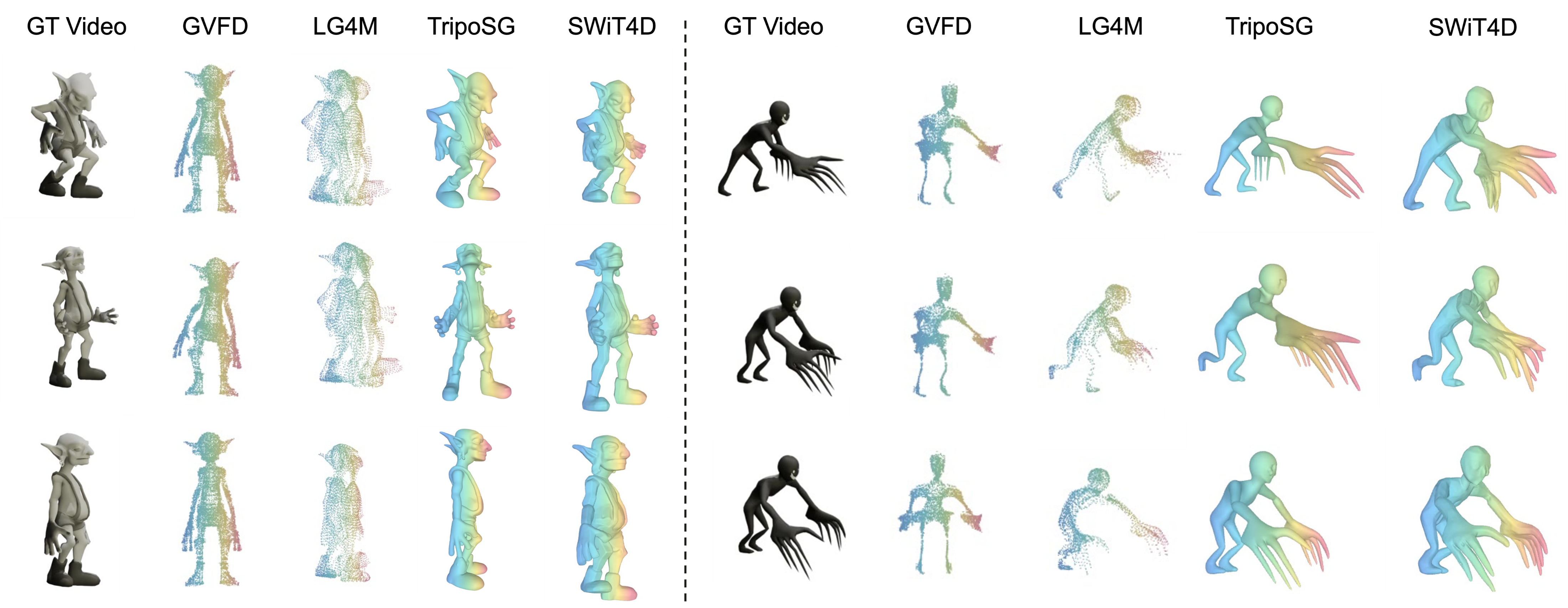}
    \caption{
    \textbf{Zero-shot generalization on unseen Objaverse assets.}
    This figure illustrate generation results on randomly sampled Objaverse assets that were \emph{not used during training}. 
    Our method generalizes robustly to diverse shapes, proportions, and articulation structures, demonstrating strong zero-shot performance on completely unseen categories.
    }
    \label{fig:objaverse-compare}
\vspace{-5pt}
\end{figure*}


\begin{table*}[t]
\centering
\small
\begin{tabular}{llcccccccc}
\toprule
& & \multicolumn{4}{c}{\textbf{Frame-Level Metrics}} & \multicolumn{4}{c}{\textbf{Temporal Metrics}} \\
\cmidrule(lr){3-6} \cmidrule(lr){7-10}
\textbf{Animal} & \textbf{Method} &
\textbf{CD} $\downarrow$ &
\textbf{F-score} $\uparrow$ &
\textbf{Precision} $\uparrow$ &
\textbf{Recall} $\uparrow$ &
\textbf{$\Delta$CD} $\downarrow$ &
\textbf{FE Cos} $\uparrow$ &
\textbf{Feat. DTW} $\downarrow$ &
\textbf{Occ. KL} $\downarrow$ \\
\midrule
Alligator & TripoSG & 0.0999 & 0.2424 & 0.2461 & 0.2401 & 0.0272 & 0.9867 & 3825.53 & 3.8411 \\
Alligator & \textbf{Ours-1shot} & \textbf{0.0194} & \textbf{0.6784} & \textbf{0.6701} & \textbf{0.6875} & \textbf{0.0014} & \textbf{0.9997} & \textbf{1498.51} & \textbf{0.3118} \\
\addlinespace[2pt]
Horse & TripoSG & 0.0544 & 0.3557 & 0.3427 & 0.3757 & 0.0401 & 0.9935 & 2047.53 & 4.2108 \\
Horse & \textbf{Ours-1shot} & \textbf{0.0182} & \textbf{0.7316} & \textbf{0.7354} & \textbf{0.7288} & \textbf{0.0011} & \textbf{0.9990} & \textbf{1133.87} & \textbf{0.2047} \\
\addlinespace[2pt]
Parrot & TripoSG & 0.0989 & 0.1918 & 0.2123 & 0.1785 & 0.0615 & 0.9962 & 1727.06 & 4.5333 \\
Parrot & \textbf{Ours-1shot} & \textbf{0.0377} & \textbf{0.4599} & \textbf{0.4606} & \textbf{0.4614} & \textbf{0.0060} & \textbf{0.9985} & \textbf{1065.06} & \textbf{0.5451} \\
\bottomrule
\end{tabular}
\caption{
Quantitative comparison between \textbf{TripoSG (baseline)} and \textbf{Ours-1shot} across different animal categories in Truebones Zoo dataset.
Metrics are divided into \textbf{Frame-Level} and \textbf{Temporal} groups.
}
\label{tab:animal_metrics_comparison}
\end{table*}


\begin{table}[]
\centering
\small
\setlength{\tabcolsep}{3pt}
\begin{tabular}{lcccc}
\toprule
\textbf{Method} &
\textbf{CD $\downarrow$} &
\textbf{F-Score $\uparrow$} &
\textbf{$\Delta$CD $\downarrow$} &
\textbf{Feat. DTW $\downarrow$} \\
\midrule
TripoSG & 0.0999 & 0.2424 & 0.0272 & 3825.53 \\
Alligator–Horse & 0.0634 & 0.2537 & 0.0027 & 3168.30 \\
Alligator-Alligator & \textbf{0.0194} & \textbf{0.6784} & \textbf{0.0014} & \textbf{1498.51} \\
\bottomrule
\end{tabular}
\caption{
Cross-species evaluation on Alligator using key metrics: Chamfer Distance (CD), F-Score, temporal Chamfer Distance change ($\Delta$CD), and Feature DTW. 
The model trained on Horse (\emph{Alligator–Horse}) improves over the TripoSG baseline when testing on Alligators, despite being cross-species. }
\label{tab:alligator_cross_species_simplified}
\end{table}

\subsection{Full-Scale and One-Shot Evaluation on In- and Out-of-Domain Benchmarks}

We begin by evaluating both \textbf{One-shot LoRA fine-tuning} and \textbf{full-scale fine-tuning} models on the Truebones Zoo (in-domain) and Objaverse (out-of-domain) test sets. These experiments are designed to highlight three key aspects: (\emph{i}) data efficiency under limited supervision, (\emph{ii}) generalization to unseen categories and mesh structures, and (\emph{iii}) temporal coherence in 4D mesh sequences.

\paragraph{In-Domain Results: Truebones Zoo.}
Table~\ref{tab:metrics-comparison-zoo} summarizes frame-level and temporal metrics on the Truebones Zoo test set. 
Our full-scale model achieves state-of-the-art performance across all metrics, with Chamfer Distance (CD) reduced by over 2.5$\times$ compared to TripoSG (0.0338 vs.\ 0.0863), and Feature DTW dropping from 1714.63 to 976.58, demonstrating both higher spatial fidelity and smoother temporal dynamics. F-score, Precision, and Recall are improved by more than twofold, and all temporal metrics ($\Delta$CD, FE Cos, Occ. KL) confirm superior motion consistency.

Even under extreme data scarcity, the \textbf{Ours-1shot} variant (trained with one single sequence) consistently outperforms all baseline methods, achieving significantly lower CD and higher F-score than the best prior models. Notably, \textbf{Ours-1shot} also yields the best or second-best results on all temporal metrics, validating the strong data efficiency and robustness of our approach.

\paragraph{Out-of-Domain Results: Objaverse.}
Table~\ref{tab:objaverse-main} shows quantitative results on the Objaverse test set, where all assets and motions are unseen during training. Our method maintains clear advantages over all baselines, achieving the lowest CD (0.0786), the highest F-score (0.2253), and the best or near-best temporal metrics. Compared with TripoSG (CD: 0.1507, F-score: 0.1112), our method more than doubles geometric accuracy and improves temporal coherence (Feat. DTW: 1729.46 vs. 1589.03, Occ. KL: 0.5616 vs. 3.6063). 

Interestingly, while variants of TripoSG with latent-space sharing or smoothing do improve temporal metrics (Feat. DTW, Occ. KL), their overall spatial and temporal quality remains substantially lower than our method, further highlighting the effectiveness of the SWiT-4D temporal mechanism.

The \textbf{Ours-1shot} variant also demonstrates strong generalization on Objaverse, consistently outperforming all baselines in both spatial and temporal measures, despite being trained on just one example per category. These results underline our model’s capacity to generalize to new shapes, topologies, and motion types with minimal supervision.

Overall, our approach demonstrates strong data efficiency, robust generalization, and high temporal consistency on both in-domain and challenging out-of-domain 4D benchmarks, outperforming existing methods in all major metrics.

\subsection{Qualitative Comparison}
\label{sec:qualitative}

We provide qualitative results to highlight the effectiveness and robustness of our method across a wide range of scenarios. For clearer and more intuitive visualization, we additionally provide an interactive demo website with full video comparisons, and we encourage readers to view the dynamic 4D results on our project page.

\noindent\textbf{In-domain (Truebones Zoo).}
As shown in Figure~\ref{fig:zoo-compare}, our method produces smoother motions, fewer artifacts, and more anatomically plausible poses compared with existing baselines.

\noindent\textbf{Out-of-domain (Objaverse).}
Figure~\ref{fig:objaverse-compare} demonstrates strong zero-shot generalization to unseen Objaverse assets, where our model preserves both geometric fidelity and temporal coherence for new shapes and articulation structures.

\begin{figure}[!t]
    \centering
    \includegraphics[width=0.45\textwidth]{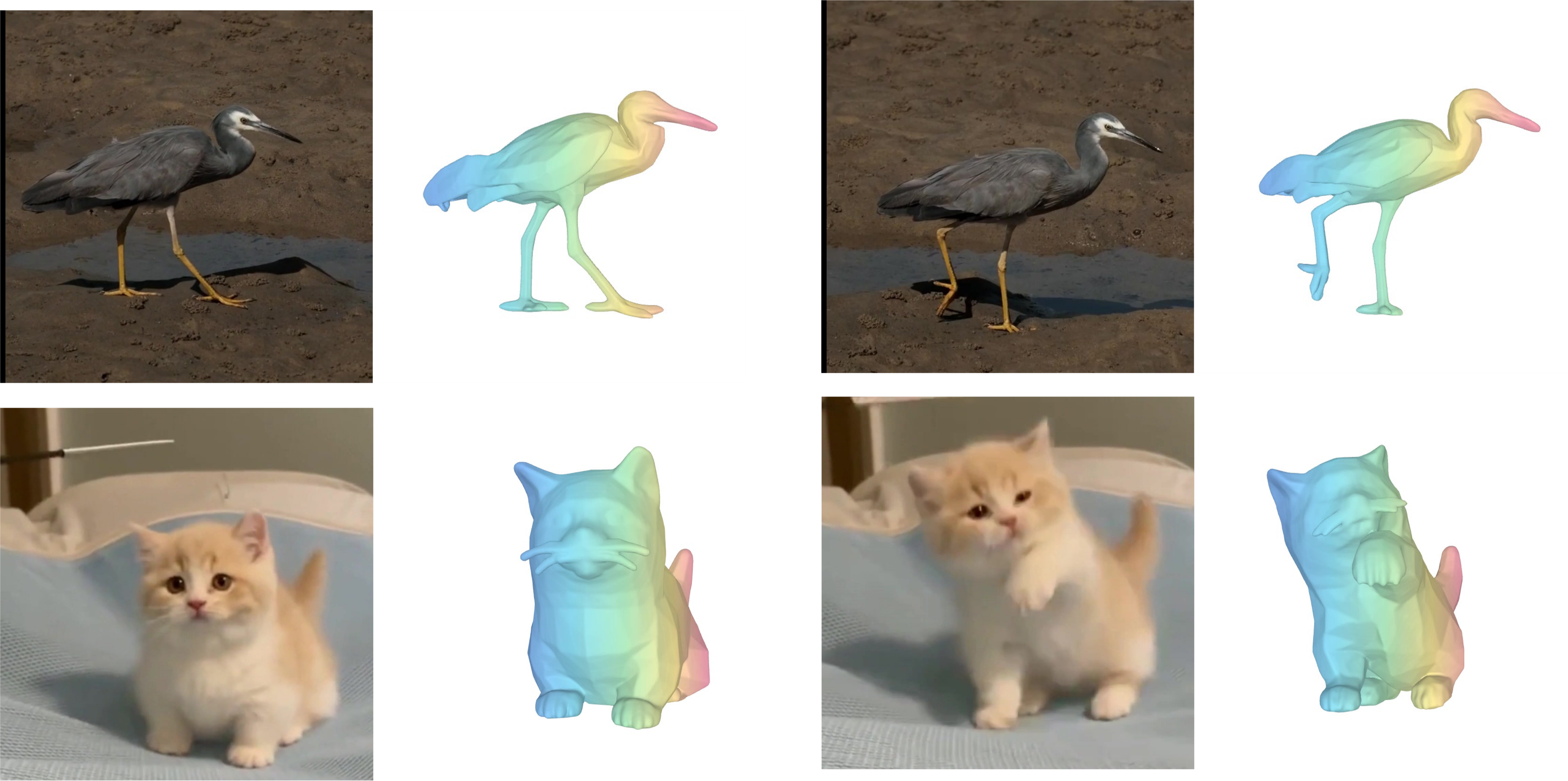}
    \caption{
    \textbf{Zero-shot generalization on unseen Consistent4D and in-the-wild videos.}
    This figure shows motion capture results on randomly sampled Consistent4D assets and challenging real-world (in-the-wild) video sequences, none of which were used during training. 
    Our method demonstrates strong robustness and generalization to novel shapes, diverse motion types, and previously unseen scenarios, producing stable and anatomically plausible 4D reconstructions in truly out-of-domain settings.
    }
    \label{fig:in-the-wild-result}
\vspace{-5pt}
\end{figure}

\noindent\textbf{Unconstrained Videos (Consistent4D and In-the-wild).}
Figure~\ref{fig:in-the-wild-result} shows that our approach remains robust under challenging real-world conditions, generating stable, coherent 4D sequences even for highly diverse and previously unseen motions.

Overall, these visual results align with our quantitative evaluations and confirm the broad applicability and strong generalization ability of our method.


\begin{figure}[!t]
    \centering
    \includegraphics[width=0.45\textwidth]{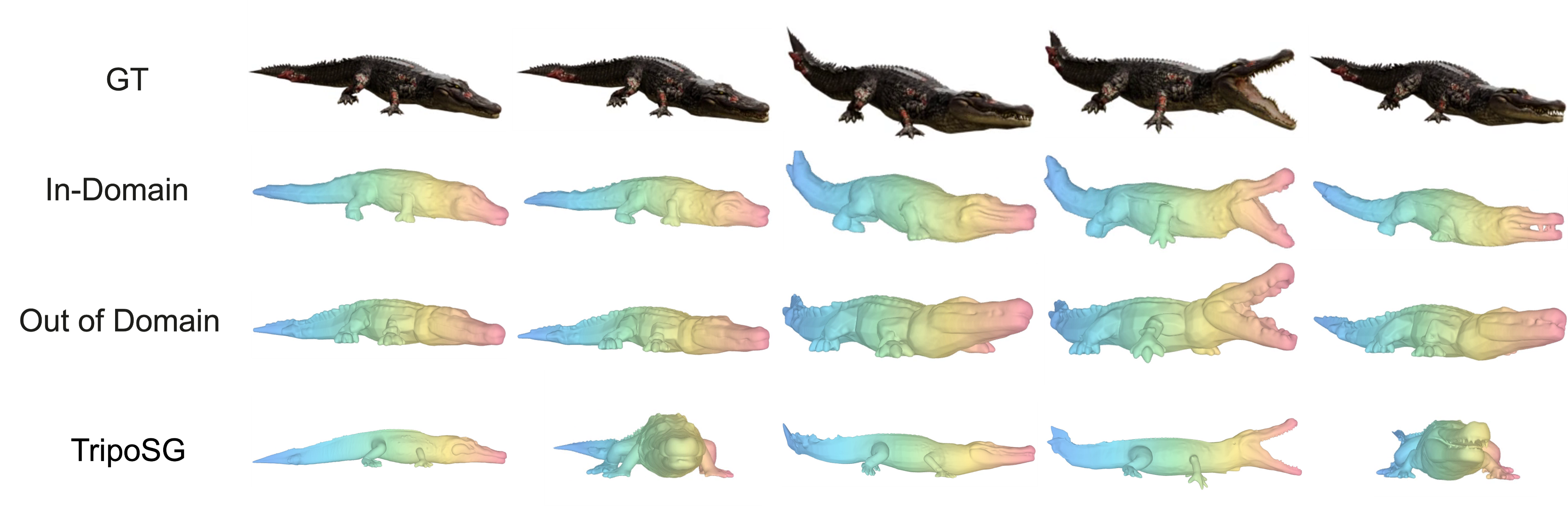}
    \caption{
    \textbf{One-shot fine-tuning results on same-species and cross-species settings.}
    These results highlight the strong adaptability and data efficiency of our framework.
    }
    \label{fig:oneshot-compare}
\vspace{-5pt}
\end{figure}


\subsection{Ablation On One-Shot Generalization}
\label{sec:oneshot-ablation}

We perform detailed ablation experiments to evaluate the impact of one-shot fine-tuning and to analyze the generalization capability of our method across both seen and unseen categories. All results are summarized in Table~\ref{tab:animal_metrics_comparison} and Table~\ref{tab:alligator_cross_species_simplified}, and visually illustrated in Figure~\ref{fig:oneshot-compare}.

\noindent\textbf{A$\rightarrow$A: One-Shot Fine-tuning on Same Species.}
We conduct one-shot LoRA fine-tuning separately on three representative species (Alligator, Horse, Parrot), and evaluate on held-out sequences of the same species. Our method shows substantial improvements over the TripoSG baseline: for instance, the F-score for Alligator increases from 0.24 to 0.68, and Chamfer Distance (CD) drops more than 5$\times$. Feature DTW also shows strong temporal smoothness gains, demonstrating that a single sequence is sufficient to greatly enhance both geometric fidelity and temporal consistency for its own category.

\noindent\textbf{B$\rightarrow$A: Cross-Species Adaptation.}
To evaluate cross-category generalization, we fine-tune the model on a sequence from one species (e.g., Horse) and test it on a different species (e.g., Alligator). As shown in Table~\ref{tab:alligator_cross_species_simplified}, the cross-species setting naturally yields lower performance than the same-species one-shot case (A$\rightarrow$A), which is expected due to substantial differences in body shape and articulation. Nevertheless, our method still achieves a clear and consistent improvement over the TripoSG baseline across all key spatial and temporal metrics. This demonstrates that the temporal mechanism transfers well even when the fine-tuning sequence comes from an unrelated species. 

\noindent\textbf{Ablation Visualization.}
Figure~\ref{fig:oneshot-compare} summarizes all ablation settings: A$\rightarrow$A (in-domain one-shot), B$\rightarrow$A (out-of-domain), and the original TripoSG baseline. Our method consistently yields the most temporally stable and semantically plausible results. We further provide a web-based demo with extensive visualizations, showing that our approach delivers dramatic improvements in temporal stability over TripoSG—regardless of whether the fine-tuned sequence is from the same or a different species.

Overall, These ablation results confirm that one-shot fine-tuning on a single sequence can greatly improve performance for both in-category and out-of-category generalization. The strong gains in temporal stability are evident in both quantitative metrics and video visualizations available on our demo page.

\section{Conclusion}
We introduced \textbf{\name}, a \emph{parameter-free} sliding-window extension that upgrades DiT-based image-to-3D backbones into video-conditioned 4D mesh generators. With 1D-RoPE over time, it guarantees \emph{lossless prior preservation} at $W{=}0$ and performs \emph{temporal residual learning} for $W{>}0$, enabling any-length inference without changing the original architecture or objectives. Instantiated on TripoSG, \name achieves high-fidelity geometry and strong temporal coherence with only limited 4D data, while remaining efficient to fine-tune. An optimization-based trajectory module further recovers accurate world-space motion under static cameras. Future work includes multi-object scenes, texture estimation and end-to-end trajectory learning under moving camera.

{
    \small
    \bibliographystyle{ieeenat_fullname}
    \bibliography{main}
}

\clearpage
\setcounter{page}{1}
\maketitlesupplementary

\section{More Visualization Results}
We provide additional qualitative visualizations to further demonstrate the performance and robustness of SWiT-4D. On the Truebones Zoo dataset, SWiT-4D generates high-fidelity 4D meshes of diverse animal species and maintains stable temporal evolution across sequences of varying lengths, producing coherent geometry and smooth motion throughout. Beyond in-domain evaluations, we also test our model on the Consistent4D benchmark to assess generalization to unseen object categories. Even without category-specific supervision, SWiT-4D reconstructs temporally consistent 4D meshes for novel shapes, indicating strong transferability across domains.

We additionally present results on challenging real-world videos. These in-the-wild reconstructions show that SWiT-4D remains robust under complex lighting, motion, and background conditions, consistently producing temporally stable and structurally coherent 4D meshes from monocular video. To evaluate data efficiency, we include a one-shot fine-tuning setting using only a single short video (approximately 150 frames). Remarkably, even with such limited supervision, SWiT-4D achieves noticeable improvements in geometric fidelity and temporal smoothness, highlighting its practical applicability under scarce 4D training data.

To recover global object motion in world coordinates, we also present reconstructions with global translation using our trajectory prediction module. Both Zoo sequences and real-world videos demonstrate accurate temporal tracking and consistent mesh trajectories.  Finally, we provide side-by-side comparisons with recent state-of-the-art approaches, including TripoSG, GVFD, LG4M, and GenZoo, on both Zoo and Objaverse datasets. Ground-truth meshes, our full method, our one-shot variant, and all baselines are visualized together. Across all evaluations, SWiT-4D produces smoother temporal dynamics, sharper geometry, and more stable reconstructions.

\section{More Experiment Results}
\begin{table}[h!]
\centering
\setlength{\tabcolsep}{3pt}
\begin{tabular}{cccccc}
\toprule
$\textbf{W}_{\text{cross}}$  & \textbf{\#SA} &
\textbf{CD $\downarrow$} &
\textbf{F-Score $\uparrow$} &
\textbf{$\Delta$CD $\downarrow$} &
\textbf{Feat. DTW $\downarrow$} \\
\midrule
C3 & Half & 0.0544 & 0.3293 & 0.0047 & 1663.31 \\
C7 & Half & 0.0383 & \textbf{0.4174} & \textbf{0.0025} & 1204.36 \\
C5 & Full & 0.0441 & 0.3847 & 0.0029 & 1296.27 \\
C5 & Half & \textbf{0.0405} & 0.4045 & 0.0029 & \textbf{1190.05} \\
\bottomrule
\end{tabular}
\caption{
\textbf{Ablation of model structures.}
We ablate the sliding window size of cross attention, varied from 3, 5, 7, as shown in C3, C5, C7 in the table. Moreover, we evaluate the performance of applying windowed attention in every self attention block(\#SA=Full), or only applied in every 2 self attention blocks(\#SA=Half). Our default setting is (C5, Half), which achieved the best trade-off over computational cost.
}
\label{tab:ablation-structure}
\end{table}

\paragraph{Ablation Study.}
To assess the influence of architectural design choices, we conduct an ablation study on two key components of our model: the sliding window size used in cross-attention layers and the frequency at which windowed self-attention is applied. As shown in Table~\ref{tab:ablation-structure}, reducing the cross-attention window to $W_{\text{cross}}=3$ negatively affects reconstruction quality and temporal consistency, while enlarging it to $7$ provides limited marginal gain with heavy computational cost. A moderate window size of $5$ consistently yields satisfying performance. Furthermore, applying windowed self-attention only in alternating layers (``Half'') outperforms applying it in every layer (``Full''), suggesting that excessive locality constraints hinder global information flow. The combination of a window size of $5$ with half-frequency self-attention achieves the best trade-off over computational cost and producing favorable results, which serves as our final architecture.

\begin{table}[h!]
\centering
\small
\begin{tabular}{lcccc}
\toprule
\textbf{Method} &
\textbf{CD $\downarrow$} &
\textbf{F-Score $\uparrow$} &
\textbf{$\Delta$CD $\downarrow$} &
\textbf{Feature DTW $\downarrow$} \\
\midrule
TripoSG & 0.0989 & 0.1918 & 0.0615 & 1727.06 \\
Parrot\_1 & 0.0377 & 0.4599 & \textbf{0.0060} & 1065.06 \\
Parrot\_3 & 0.0340 & 0.4926 & 0.0066 & 985.79 \\
Parrot\_6 & \textbf{0.0305} & \textbf{0.5451} & 0.0066 & \textbf{943.24} \\
\bottomrule
\end{tabular}
\caption{
Quantitative comparison of \textbf{Parrot} across different checkpoints.
TripoSG denotes the baseline model, while higher indices (1, 3, 6) indicate progressively improved variants.
Arrows indicate whether higher ($\uparrow$) or lower ($\downarrow$) values are better.
CD: Chamfer Distance.
}
\label{tab:parrot-metrics-progression}
\end{table}

Tab.~\ref{tab:parrot-metrics-progression} illustrates how temporal coherence evolves as the number of Parrot training sequences increases. We quantify temporal smoothness using a Dynamic Time Warping (DTW) score computed on latent motion features, where lower values correspond to better frame-to-frame consistency. Starting from the TripoSG baseline, the temporal alignment improves progressively as the model is trained with 1, 3, and 6 Parrot sequences. This monotonic reduction in Feature DTW indicates that additional Parrot supervision enables the model to generate motions with smoother dynamics and more stable temporal structure.

\section{Implementation Details}

\subsection{Dataset Details}

\paragraph{One-Shot Fine-Tuning Clip Selection.}
For the one-shot setting, we randomly select a single sequence that contains approximately 150 contiguous frames. No additional heuristics or motion filtering are applied. The selected clip is used exclusively for fine-tuning, while another 5 random sequences from the same species are sampled as test data.

\paragraph{Mesh Normalization and Preprocessing.}
Each mesh sequence undergoes a two-stage normalization procedure to ensure consistent scale and canonicalization across videos and species.
First, we compute the bounding box of the rest-pose mesh for each identity and scale all meshes by this box so that the rest pose fits inside a unit cube.
Next, for every frame, we remove the global translation by centering the mesh at the origin.
We then compute a super bounding box covering all frames in the sequence and uniformly scale the entire sequence to lie within the range $\left[-1, 1\right]^3$.
This normalization ensures a consistent spatial scale for training and evaluation.

\paragraph{In-the-Wild Videos.}
For in-the-wild inputs, we assume a fixed external camera throughout the entire video. 
No camera pose optimization or multi-view geometric processing is performed under this setting.

\subsection{Experimental Settings}
We use a consistent set of hyperparameters for one-shot and few-shot adaptation across all species, training with Adam optimizer, a learning rate of $1\times10^{-4}$, batch size of $1$ per NPU, a temporal window of $48$ frames, and a hopping stride of $24$ frames. Full-scale Fine-tuning proceeds for $18{,}000$ optimization steps, while One-shot LoRA Fine-
tuning takes $2{,}000$ steps. The model is initialized from the TripoSG pretrained checkpoint trained on Objaverse. Fine-tuning is performed solely on the Truebones Zoo dataset, while evaluation is conducted on both Truebones Zoo and Objaverse to assess cross-dataset generalization. Experiments are run on a compute cluster with 8 NPUS and 64\,GB system memory, and the training requires approximately two days (8 NPUs) and 4 hours for One-shot LoRA (2 NPUs).

\subsection{Evaluation Metrics}

\paragraph{Frame-level Metrics.}
To evaluate the geometric accuracy of each generated frame, we compute 
\emph{Chamfer Distance}, \emph{Precision}, \emph{Recall}, and \emph{F-Score} 
between the predicted mesh and the ground-truth mesh. 
Following the implementation used in our experiments, 
all metrics are computed on point clouds sampled uniformly from the mesh surface.

\paragraph{Surface Point Sampling.}
Given a mesh with vertices $V=\{v_i\}_{i=1}^N$ and triangular faces 
$F=\{(i,j,k)\}$, we uniformly sample $n$ surface points. 
For each triangle $t=(v_i,v_j,v_k)$, we compute its area
\[
A_t = \frac{1}{2}\left\| (v_j - v_i) \times (v_k - v_i) \right\|
\]
and draw triangles proportionally to their areas.  
Each sampled point is generated via barycentric sampling:
\[
p = (1-\sqrt{r_1})\, v_i 
    + \sqrt{r_1}(1-r_2)\, v_j 
    + \sqrt{r_1} r_2\, v_k,
\]
where $r_1,r_2 \sim \mathcal{U}(0,1)$.

Let 
\[
P = \{p_i\}_{i=1}^n \quad \text{and} \quad 
G = \{g_i\}_{i=1}^n
\]
denote sampled point sets from the predicted mesh and the ground-truth mesh, respectively.

\paragraph{Nearest-Neighbor Distances.}
For each point $p \in P$, we compute the nearest Euclidean distance to $G$:
\[
d_{P\to G}(p) = \min_{g\in G}\|p-g\|_2,
\]
and similarly for $g \in G$:
\[
d_{G\to P}(g) = \min_{p\in P}\|g-p\|_2.
\]

\paragraph{Chamfer Distance.}
We compute the (non-squared) bi-directional Chamfer Distance:
\[
\mathrm{CD} 
= \frac{1}{2}\left(
    \frac{1}{|P|} \sum_{p\in P} d_{P\to G}(p)
    +
    \frac{1}{|G|} \sum_{g\in G} d_{G\to P}(g)
  \right).
\]

\paragraph{Precision, Recall, and F-Score.}
Given a threshold $\tau$ (set to $0.02$ in our experiments), 
we define precision as the fraction of predicted points within $\tau$ of the ground truth:
\[
\mathrm{Precision} 
= 
\frac{1}{|P|}
\sum_{p\in P} \mathbb{1}\!\left[ d_{P\to G}(p) < \tau \right].
\]
Recall measures the fraction of ground-truth points recovered by the prediction:
\[
\mathrm{Recall} 
= 
\frac{1}{|G|}
\sum_{g\in G} \mathbb{1}\!\left[ d_{G\to P}(g) < \tau \right].
\]
The F-score combines both terms:
\[
\mathrm{F\!-\!Score}
= 
\begin{cases}
\dfrac{2\,\mathrm{Precision}\cdot\mathrm{Recall}}
      {\mathrm{Precision}+\mathrm{Recall}},
& \text{if Precision + Recall} > 0, \\
0, & \text{otherwise}.
\end{cases}
\]

\paragraph{Temporal Chamfer Delta ($\Delta$CD).}
While frame-level Chamfer Distance measures static geometric accuracy, it does not 
capture whether the \emph{motion pattern} of the generated sequence matches the 
ground-truth motion.  
To quantify temporal evolution, we compute a Chamfer Distance for every 
pair of consecutive frames, both for the prediction and the ground truth.

Let $P_t$ and $G_t$ denote the point clouds sampled from the predicted and 
ground-truth meshes at frame $t$.  
The frame-to-frame Chamfer Distance for prediction is

\[
\begin{aligned}
\mathrm{CD}^{\mathrm{pred}}_t
&=
\frac{1}{2}\Big(
    \frac{1}{|P_t|} \sum_{p\in P_t} 
        \min_{q\in P_{t+1}}\|p-q\|_2
    +
\\[-2pt]
&\qquad
    \frac{1}{|P_{t+1}|} \sum_{q\in P_{t+1}} 
        \min_{p\in P_t}\|p-q\|_2
\Big),
\end{aligned}
\]

and similarly for ground truth:

\[
\begin{aligned}
\mathrm{CD}^{\mathrm{gt}}_t
&=
\frac{1}{2}\Big(
    \frac{1}{|G_t|} \sum_{g\in G_t} 
        \min_{h\in G_{t+1}}\|g-h\|_2
    +
\\[-2pt]
&\qquad
    \frac{1}{|G_{t+1}|} \sum_{h\in G_{t+1}} 
        \min_{g\in G_t}\|g-h\|_2
\Big).
\end{aligned}
\]

The temporal Chamfer Delta measures how similarly the two sequences evolve:
\[
\Delta\mathrm{CD}
=
\frac{1}{T-1}
\sum_{t=1}^{T-1}
\left|
\mathrm{CD}^{\mathrm{pred}}_t - \mathrm{CD}^{\mathrm{gt}}_t
\right|.
\]

A small $\Delta$CD indicates that the predicted motion exhibits a 
temporal deformation pattern similar to the ground truth, 
even if the per-frame geometry is imperfect.  
This metric therefore complements static CD by capturing 
\emph{temporal smoothness and motion fidelity}.
\vspace{6pt}

\paragraph{Temporal Occupancy KL.}
To assess temporal consistency from a volumetric perspective, 
we further compare \emph{voxel occupancy transitions} across frames.  
For each sequence, we first determine a global bounding box 
covering all predicted and ground-truth frames.  
Given a grid resolution $K$ (we use $K=32$), the space is discretized 
into $K^3$ voxels.

For each frame $t$, the point cloud $X_t$ is voxelized into an 
occupancy histogram $o_t \in \mathbb{R}^{K^3}$:
\[
o_t(i) = \text{\#points falling into voxel $i$},
\quad
i = 1,\dots,K^3.
\]
We convert $o_t$ into a probability distribution:
\[
p_t = \frac{o_t + \varepsilon}{\sum_i (o_t(i) + \varepsilon)}, \qquad
\varepsilon = 10^{-8}.
\]

To capture how occupancy evolves from $t$ to $t+1$, we compute the 
Kullback--Leibler divergence
\[
\mathrm{KL}^{\mathrm{gt}}_t 
= 
\mathrm{KL}(p^{\mathrm{gt}}_{t+1} \,\|\, p^{\mathrm{gt}}_t)
=
\sum_{i=1}^{K^3}
p^{\mathrm{gt}}_{t+1}(i)\,
\log
\frac{
p^{\mathrm{gt}}_{t+1}(i)
}{
p^{\mathrm{gt}}_t(i)
},
\]
and similarly for the prediction:
\[
\mathrm{KL}^{\mathrm{pred}}_t 
= 
\mathrm{KL}(p^{\mathrm{pred}}_{t+1} \,\|\, p^{\mathrm{pred}}_t).
\]

The final temporal occupancy KL metric is the mean absolute discrepancy:
\[
\Delta\mathrm{KL}_{\mathrm{occ}}
=
\frac{1}{T-1}
\sum_{t=1}^{T-1}
\left|
\mathrm{KL}^{\mathrm{pred}}_t
-
\mathrm{KL}^{\mathrm{gt}}_t
\right|.
\]

This metric captures whether the \emph{volumetric shape evolution} of the generated 
meshes matches that of the ground truth.  
It is complementary to $\Delta$CD, since it is sensitive not only to surface 
deformation but also to changes in interior occupancy patterns, providing a 
richer description of 4D motion consistency.

\paragraph{Temporal Feature Embedding Metrics.}
To evaluate temporal consistency at a high-level semantic level, we extract 
per-frame features using a pretrained PointNet++ encoder 
(SSG variant without normals).  
Given a mesh sequence $\{M_t\}_{t=1}^T$, we uniformly sample point clouds 
$X_t \in \mathbb{R}^{N\times 3}$ from each mesh and feed them through the encoder:
\[
f_t = E(X_t) \in \mathbb{R}^D,
\]
where $D=1024$ in our setting.  
This yields a temporal feature sequence
\[
F = (f_1, f_2, \dots, f_T) \in \mathbb{R}^{T\times D}.
\]

We compare two mesh sequences A and B via their feature sequences
\[
F^{(A)} = (f^{(A)}_1, \dots, f^{(A)}_T), \qquad
F^{(B)} = (f^{(B)}_1, \dots, f^{(B)}_T).
\]

\vspace{6pt}
\paragraph{Feature Cosine Similarity.}
To measure global semantic similarity of two motion sequences, 
we compute cosine similarity between the \emph{mean-pooled features}:
\[
\bar{f}^{(A)} = \frac{1}{T}\sum_{t=1}^T f^{(A)}_t, 
\qquad
\bar{f}^{(B)} = \frac{1}{T}\sum_{t=1}^T f^{(B)}_t.
\]
The cosine similarity is
\[
\mathrm{Cosine}(A,B)
=
\frac{
\bar{f}^{(A)} \cdot \bar{f}^{(B)}
}{
\|\bar{f}^{(A)}\|_2 \;\|\bar{f}^{(B)}\|_2
}.
\]
A value near 1 indicates high semantic agreement between the two motion sequences.

\vspace{6pt}
\paragraph{Feature Dynamic Time Warping (DTW).}
To compare temporal evolution independent of frame rate or misalignment, 
we compute Dynamic Time Warping (DTW) on the per-frame feature trajectories.
Let
\[
d_{ij} = \left\| f^{(A)}_i - f^{(B)}_j \right\|_2
\]
be the pairwise feature distance matrix.  
DTW computes the minimal cumulative alignment cost via dynamic programming:
\[
D(i,j) 
= d_{ij}
+ \min\big( D(i-1,j),\; D(i,j-1),\; D(i-1,j-1) \big),
\]
with boundary conditions
\[
D(0,0)=0, \qquad 
D(i,0)=D(0,j)=+\infty \ \text{for } i,j>0.
\]
The resulting DTW feature distance is
\[
\mathrm{DTW}(A,B) = D(T,T).
\]

Lower DTW indicates better temporal alignment in the learned feature space, 
capturing semantic motion similarity beyond raw geometry.

\end{document}